# Bi-directional memory-based dialog translation: The KEMDT approach *

Geunbae Lee  and  Hanmin Jung† and  Jong-Hyeok Lee

Department of Computer Science and Engineering
Pohang University of Science and Technology
San 31, Hyoja-Dong, Pohang, 790-784, Korea
Tel: +82-562-279-2254, Fax: +82-562-279-2299
E-mail: gblee@vision.postech.ac.kr

## Abstract

A bi-directional Korean/English dialog translation system is designed and implemented using the memory-based translation technique. The system KEMDT (Korean/English Memory-based Dialog Translation system) can perform Korean to English, and English to Korean translation using unified memory network and extended marker passing algorithm. We resolve the word order variation and frequent word omission problems in Korean by classifying the concept sequence element in four different types and extending the marker- passing-based-translation algorithm. Unlike the previous memory-based translation systems, the KEMDT system develops the bilingual memory network and the unified bi-directional marker passing translation algorithm. For efficient language specific processing, we separate the morphological processors from the memory-based translator. The KEMDT technology provides a hierarchical memory network and an efficient marker-based control for the recent example-based MT paradigm.

## 1 Introduction

The role of memory-based parsing in natural language processing and machine translation is recently increasing due to the limitations of rule-based mechanism. In memory-based parsing, initiated by direct memory access parsing (DMAP) [Riesbeck and Martin, 1985], the parsing is viewed as a memory search problem using parallel marker passing algorithm for making inferences [Norvig, 1989; Yu and Simmons, 1990]. This approach is contrasted with the classical rule-based parsing [Dyer, 1983], and is conceptually similar to the related natural language processing techniques such as case-based parsing [Riesbeck and Schank, 1989] and example-based machine translation [Nagao, 1984]. The memory-based parsing is also applied to the machine translation problem [Kitano, 1991; Tomabechi, 1987], and several parallel computers such as SNAP and IXM2 are used to implement the memory-based machine translation systems [Kitano and Higuchi, 1991; Kitano et al., 1991].

Memory-based machine translation uses a set of

*This research was supported in part by KOSEF and Korea Research Foundation (non-directed research fund).
†Current address: Natural Language Processing Lab., Systems Engineering Research Institute (SERI/KIST), E-eun-dong, Yu-sung-ku, Taejon, Korea, Tel: +82-42-869-1456

linguistic patterns, called concept sequences (CS's) and concept nodes (CN's), organized as hierarchical structures in the semantic network based knowledge representation called memory network [Kitano, 1990; 1991]. The translation is performed, often in parallel, through linguistic pattern matching and instantiation on the memory network. In this paper, we describe a Korean/English memory-based dialog translation (KEMDT) system. We design a memory network and an extended marker passing algorithm to process the idiosyncratic features of Korean such as partially free word order, frequent word omissions, and post-positional case marking. The memory-based machine translation is especially suitable to the bi-directional dialog translation in a restricted domain because the memory network is inherently bilingual by containing source and target language's linguistic features together [Kitano, 1991], and the bi-directional dialog translation in a restricted domain is the main target of the current speech translation researches [Morimoto and Kurematsu, 1993]. There are practical reasons for developing a bi-directional dialog translation model. The dialog translation should always be at least bilingual because dialog needs at least two partners communicating each other. In the memory-based translation model, the same memory network can be shared in either translation direction, and the model is very promising when we consider the multi-lingual translation extension [Funaki, 1993]. In this paper, we will show how the common memory network design can simultaneously be utilized in English-Korean and Korean-English bi- directional dialog translation in a restricted travel domain. Also we develop an extended marker passing algorithm for bilingual translation in the common memory network.

This paper is organized as follows. In section 2, we briefly describe the linguistic characteristics of Korean. In section 3, we explain the new memory network design for handling idiosyncrasies of Korean and for bilingual translation, and also we describe the extended marker passing algorithm for parsing and generation. In section 4, we develop the architecture of our prototype translator and in section 5, we compare and contrast our model to the previous researches. Section 6 discusses the implementation and limitations, and finally section 7 draws some conclusions.

## 2 Features of Korean language

Korean, which can be classified into a morphologically agglutinative and syntactically SOV languages, has several unique linguistic features. In this paper, the Yale romanization is used for representing the Korean phonemes.

1) A Korean word, called Eojeol, consists of more than one morphemes with clear-cut boundaries in between. For example, an Eojeol *pha-il+tul+ul (files [obj])* consists of 3 morphemes:

   pha-il (file) + tul (plural suffix) + ul (object case-marker)

2) Korean is a postpositional language with many kinds of noun-endings, verb-endings, and prefinal verb-endings. These functional morphemes determine the noun's case roles, verb's tenses, modals, and modification relations between Eojeols. For example, in *swu-ceng-ha+yess+ten pha-il (the file that was edited)*, the verb *swu-ceng-ha (edit)* is of past tense and modifies *pha-il (file)* according to the given verb-endings:

   swu-ceng-ha (edit) + yess (past tense pre-final verb-ending) + ten (adnominal verb-ending)

3) Korean has relatively free word order compared to SVO languages, such as English, except for the very rigid constraints that the verb must appear in a sentence-final position, and the major syntactic components can be frequently omitted.

4) In Korean, there are some word-order constraints such that the auxiliary verbs representing modalities must follow the main verb, and the modifiers must be placed before the word (called head) they modify.

These linguistic features of Korean should be considered when we design the memory network and the translation algorithm. In the next section, we explain how we extend the conventional memory network and translation algorithm to cope with the linguistic peculiarities of Korean.

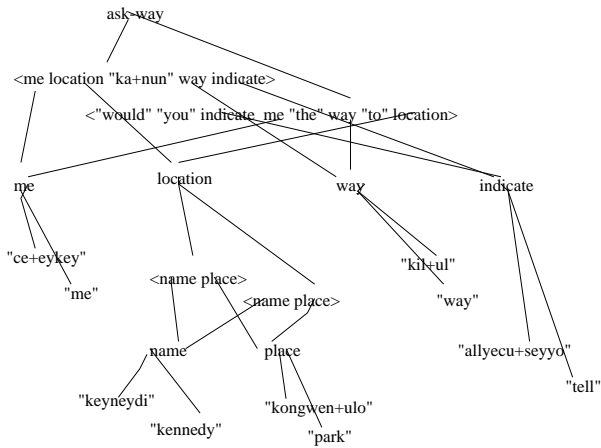

Figure 1: Example memory network for bi-directional translation of "Would you tell me the way to Kennedy Park?".

## 3 Bilingual memory network and extended marker passing for Korean

The memory network plays a linguistic knowledge base in a memory-based translation system and generally consists of concept nodes (CN's) and concept sequences (CS's) which are structured in IS-A hierarchies. The CN designates objects and events in a natural language description and the CS designates the sequential constraints in the natural language which corresponds to the syntax information in the phrase structure grammars. For bi-directional translation, the memory network must be bilingual and must have Korean CS and English CS. Figure 1 shows the example bilingual memory network which is manually collected from the Korean/English bilingual corpus. Each CN has Korean CS and English CS pairs which can be different in length (number of elements). The CS consists of the concept sequence elements (CSE's) each of which is connected to the corresponding CN through the IS- A hierarchy. The CN can be shared between Korean CS and English CS. The end of IS-A hierarchy is the Korean lexical item and the English lexical item pairs in a morphologically segmented form.

To handle the word order variation and frequent word omission in Korean, we classify the CSE into four different types: compulsory and fixed order CSE (CX), compulsory and free order CSE (CF), omissible and fixed order CSE (OX), and omissible and free order CSE (OF). For example, the following two Korean CSs show the different CSE types.

- KCS1: [me(OF) location(CX) "kanun"(CX) way(CX) indicate(CX)]

- KCS2: [me(OF) where(CF) location(CX) "issnunci"(CX) indicate(CX)]

The KCS1 can provide the syntactic structure for the Korean sentences that can be translated into the English sentence "Would you tell me the way to Kennedy Park", and the KCS2 provide the structure for the English sentence "Would you tell me where the Kennedy Park is". Each CSE type controls the marker passing which will be described shortly. In English, every CSE is classified as CX type.

We employ four different markers for parsing and generation: AA (analysis activation), AP (analysis prediction), GA (generation activation), and GP (generation prediction) markers. The markers are generally information carrying tokens that pass through the memory network to make activations and predictions. In our system, we use the simplest bit markers for checking the activation and prediction status. The functions of the four makers are similar to be defined in other literature [Riesbeck and Martin, 1985; Kitano, 1990]:

- AA-markers, which designate node activation, and passed up the IS-A hierarchy.

- AP-markers, which are passed through the source language CSE to predict the next CSE, and also passed down the IS-A hierarchy for lexical prediction.

- GA-markers, which designate node generation, and passed up the IS-A hierarchy in the target language side.

- GP-markers, which are passed through the target language CSE to predict the next generated CSE.

The basic marker passing algorithm analyzes the input sentences by repeating the predict-activate-collision cycles after the initial prediction. The initial prediction puts the AP-markers for all the first CSEs in the source language CS, and the AP-markers are passed down the hierarchy to the lexical items. The activate cycle puts the AA markers in the input lexical items. The collision cycle handles the AP - AA collisions, and passed the AA markers up the IS-A hierarchy and the collision in the CSE passes the AP markers to the next CSE in the CS. This AP-AA collision corresponds to the "shift" in the ordinary shift-reduce parsing. The prediction cycle passes the AP markers in the next CSE to down the hierarchy to the lexical items. When the collision occurs at the end of the CSE in the CS, the entire CS is accepted and the AA markers are passed to the higher level CNs, which corresponds to the "reduce" in the shift-reduce parsing. The generation is simultaneously performed with the analysis [Kitano, 1990] and used the same predict-activate-collision cycles with the GA and GP markers. The predict cycle puts the GP-markers for all CSEs in the target language CS. When the AA and AP collision occurs at the CN in the analysis phase, the activation cycle makes the GA markers passed up to the corresponding target language CSE where the GP-GA collision occurs. The GP-GA collision finally lexicalizes the CSE with the proper order. However, the original marker passing algorithm was developed for the rather rigid word order languages such as English, where the prediction sequence can be definitely determined in advance. For the partially free word order languages such as Korean, we have to extend the prediction mechanism to cover the free word order (such as CF, OF types) and omissible CSE types (such as OF, OX types) in the analysis and generation.

Figure 2 shows the marker movements for the free order (CF, OF types, figure 2-a) and omissible fixed order (OX type, figure 2-b) CSEs. In figure 2-a), the free order CSE must always be predicted. When the AA-marker meets the AP-marker in the free order CSE, the CSE is accepted in that position. Otherwise, the free order CSE must wait for the AA-marker while other CSEs are accepted. The figure 2-b) shows the marker movement on the omissible fixed order CSEs. The CSEs must be predicted together with the next CSE since it can be omitted. If the AA-marker hits the next CSE, then the OX CSE's prediction will disappear automatically even if it is not activated at all. In the generation, the extended marker movements are the same except that the GA-markers and GP-markers are used instead of AA-markers and AP-markers.

The design of memory network and the extended marker passing focuses on the bi-directionality of the translation, and therefore the extended marker passing provides uniform translation mechanism from Korean to English and vice versa.

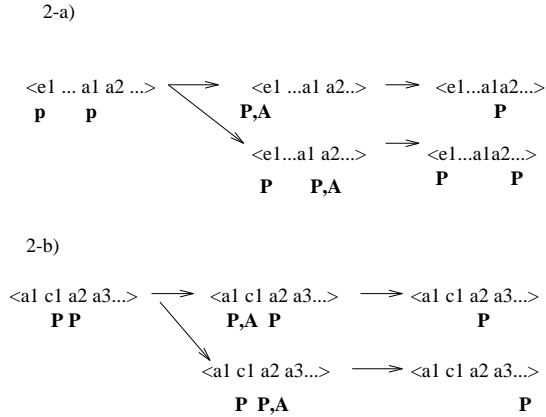

Figure 2: Extended prediction in the analysis (P: AP-marker, ai:compulsory and fixed order CSE, ei: compulsory and omissible free order CSE, ci: omissible fixed order CSE).

## 4 Architecture of the KEMDT system

Figure 3 shows the overall architecture of the KEMDT system. The bilingual memory network and the marker passing mechanism are as explained in the previous section. The Korean morphological processing is implemented using the CYK morphological analyzer [Lee and Lee, 1992 in Korean], and the English processing is performed using the two-level morphological analyzer [Koskenniemi, 1983]. The morphological processing handles both morphological analysis and generation, and also handles the irregular conjugations in Korean and English. For example, "study+s" is generated as "studies" in English and "kop+un" is generated as "kowun" (beautiful) in Korean.

The bi-directional translation between Korean and English is performed through the extended marker passing and generally follow the procedures:

1. initial prediction: put the AP-makers on all the first CSEs in each CS, and send them down to the lexical items.

2. morphological processing: morphologically segment the input words and match them with the lexical items. If the matching fails, then perform the spelling correction (which is not explained in this paper) and try to match the corrected words again. For generation, the lexical items in the network are generated into the proper surface forms (handling irregular conjugation (or inflection)).

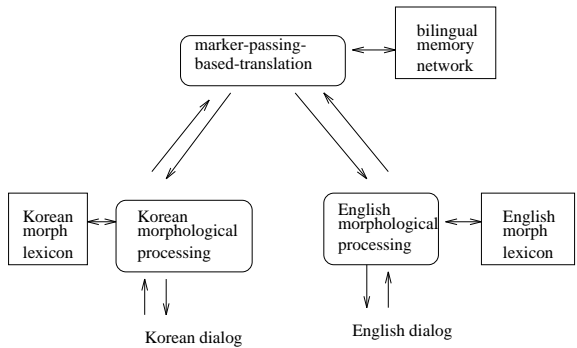

Figure 3: KEMDT system architecture. The ovals are processing modules and the squares are memory modules. The arrows designate the data flows in the system. Note the system performs bi-directional translation without any extra system modules.

3. concept activation & propagation: put the AA-markers in the matched lexical items. If the lexical item has the AP-markers, the AP-AA collision occurs, which 1) makes the corresponding GA- markers put on the target language's lexical items, and 2) sends the AA-markers up the IS-A hierarchy. The GA-markers are also propagated to the target language CS through the IS-A hierarchy.

4. prediction update: The AA-AP collision in the source language CS moves the AP-marker to the next CSE. The GA-GP collision in the target language CS also moves the GP-marker to the next CSE. The CSE is morphologically generated when the GA-GP collision occurs. However, in the special case, the lexical item CSE is directly generated with only GP-markers when there is no corresponding source language lexical items. Steps 3 - 4 are repeated until the entire CS is accepted.

5. CS accept: When the collision occurs at the last CSE, the CS is accepted and the whole target language CS is morphologically generated.

## 5 Comparison to the previous systems

KEMDT is a memory-based translation system which is different from the classical rule-based (transfer and interlingua) machine translation systems. The rule-based MT performs deep linguistic analysis using abstract and condensed linguistic rules, and translates according to the structure transfer or interlingua management. The memory-based MT performs a simple pattern match against the built-in memory network for the necessary translation. The memory-based MT naturally supports the bi- directional translation because the linguistic knowledge base usually contains both source and target language's linguistic information, which is difficult for the classical rule-based MT.

Many features of KEMDT system design are firstly influenced by Dm- Dialog, the first speech to speech memory-based MT system [Kitano, 1990; 1991]. However, it is unclear how Dm-Dialog can handle the bi-directionality in translation in uniform memory network and uniform marker passing technique. Moreover, the Dm-Dialog technique cannot be directly applied to the Korean language, which is very different in linguistic typology from English, especially in word order variation and frequent word omission. The KEMDT system also differs from the Dm-Dialog in having separate morphological processing modules for more efficient language specific processing that cannot be done by the marker passing alone. The KEMDT memory network is designed using only CN's and CS's, which provides clear memory structures for non experts to design and extend the necessary memory networks. The lexical items are in morphologically analyzed forms for providing linguistic generality.

The recent example-based MT researches give promising results for the case-based and memory-based approaches for the machine translation. The KEMDT design is also influenced by the recent example-based MT researches [Sato and Nagao, 1990; Sumita and Iida, 1991; Kitano, 1993]. These systems use aligned bilingual translation examples for selecting, modifying and generating translations. The bi-lingual example-base is similar to the KEMDT's bilingual memory network, and it is a still open problem to decide whether the phrase level or sentence level or text level bilingual example-bases are efficient. The example- based approaches are still mainly used for the source to target transfers, but rarely used for the full MT systems yet[1] The KEMDT system has a memory network which is considered to be a hierarchically structured example-base and this hierarchy makes the simple translation-control using marker passing possible and feasible. Moreover, the bi-directional translation is performed using the uniform marker

---
[1] One notable exception is TDMT approach [Furuse and Iida, 1992].

passing technique, which has not been tried in the previous example-based MT systems yet.

## 6 Implementation and limitations

The KEMDT system is implemented on a unix workstation using C and Motif user interface. The user interface windows consist of the memory network, instance hierarchy for the translation, current input sentence, and translation result history respectively. Currently the system performs the translation for dialogs that are usually exchanged in foreign country travel. The memory network currently contains about thousands of Korean and English words, and a few hundreds of concepts and concept sequences, which are manually constructed from the Korean/English bilingual corpus.

The current system can evolve to multi-lingual MT systems by designing multi-lingual memory networks, and modifying marker passing algorithms to process multi-lingual translation. Moreover, the system can be extended to speech to speech MT systems by incorporating the currently developing integrated speech and natural language morphological processing techniques [Kim et al., 1994]. However, there are some limitations to the current KEMDT approach. First, the example memory network, although it is much more simplified than the original memory networks used in [Kitano, 1990], still must be constructed by the conventional knowledge encoding method (for IS-A hierarchy and the concept sequence) which hinders the practical scaling up. We are searching the ways to semi-automatically construct the memory networks using the bilingual corpus. Secondly, in activating the lexical items, the current implementation just uses exact matches which is obviously limited. However, implementing the best matches using the Korean and English thesaurus [Sato and Nagao, 1990; Sumita and Iida, 1991] is not difficult in the current model, and we are implementing the best match scheme now. Finally, the system still runs on the sequential machine (like SUN sparc). We have to implement the extended marker passing algorithm on the parallel machines such as [Kitano and Higuchi, 1991; Kitano et al., 1991] to fully exploit the parallelism inherent in the marker passing.

## 7 Conclusion

The memory-based bi-directional dialog translation system between English and Korean is designed and implemented. The system KEMDT contributes to the memory-based MT researches by proposing 1) bilingual memory network and uniform marker passing for bilingual translation, 2) efficient handling of word order variation and frequent omissions which are ubiquitous in the agglutinative languages such as Korean in memory-based framework, and 3) integrated morphological processing with the memory-based paradigm. We plan to develop a multi-lingual dialog translation system based on the memory-based, uniform marker passing technology in the future.